\documentclass[ba]{imsart}
\pubyear{2022}
\volume{TBA}
\issue{TBA}
\arxiv{1903.09668}
\firstpage{1}
\lastpage{1}

\usepackage{amsthm}
\usepackage{amsmath}
\usepackage{graphicx}
\usepackage{enumerate}
\usepackage{natbib}
\usepackage{url} 
\usepackage{float}
\usepackage{booktabs}

\usepackage{enumitem}

\usepackage{algorithm}
\usepackage[noend]{algpseudocode}
\usepackage{multirow}
\usepackage{lipsum}
\usepackage{amsfonts}
\usepackage{cite}      

\usepackage{amsfonts}
\usepackage{amssymb}
\usepackage{subfigure}
\usepackage{multirow}
\usepackage{color}
\usepackage{blkarray}

\usepackage[dvipsnames]{xcolor}
\usepackage[colorlinks,citecolor=blue,urlcolor=blue,filecolor=blue,backref=page]{hyperref}
\usepackage{cleveref}

\newcommand{\bm}[1]{\boldsymbol{#1}}

\renewcommand{\d}{\mathrm{d}}

\newcommand{\E}{\mathbb{E}}

\def\pg{{\mathcal{PG}}}
\def\gig{{\mathcal{GIG}}}
\def\mE{{\mathcal{E}}}

\newcommand{\Var}{\mathsf{Var\,}}
\newcommand{\iid}{\stackrel{iid}{\sim}}

\newcommand{\diag}{\text{diag}}

\newcommand{\norm}[1]{\left\Vert#1\right\Vert}
\newcommand{\abs}[1]{\left\vert#1\right\vert}

\def\bomega{\bm{\omega}}

\def\B{\bm{B}}

\def\x{\bm{x}}
\def\y{\bm{y}}

\def\btheta{\bm{\theta}}
\def\bTheta{\bm{\Theta}}
\def\Var{\text{Var}}
\def\Cov{\text{Cov}}
\renewcommand{\d}{\mathrm{d}\,}

\def\mN{\mathcal{N}}

\def\E{\qopname\relax o{E}}

\newcommand{\bx}{{\bm x}}

\newcommand{\by}{{\bm y}}
\newcommand{\bz}{{\bm z}}
\newcommand{\bepsilon}{{\bm \epsilon}}
\newcommand{\blambda}{{\bm \lambda}}

\newcommand{\R}{\mathbb{R}}
\newcommand{\defeq}{\mathrel{\mathop:}=}

\startlocaldefs
\numberwithin{equation}{section}
\theoremstyle{plain}

\newtheorem{lemma}{Lemma}[section]
\newtheorem{remark}{Remark}[section]
\endlocaldefs

\begin{document}

\begin{frontmatter}
\title{Data Augmentation for  Bayesian  Deep Learning}
\runtitle{Data Augmentation for Bayesian   Deep Learning}

\begin{aug}
\author{\fnms{Yuexi} \snm{Wang}\thanksref{addr1}\ead[label=e1]{yuexi.wang@chicagobooth.edu}},
\author{\fnms{Nicholas} \snm{Polson}\thanksref{addr1}\ead[label=e2]{ngp@chicagobooth.edu}}
\and
\author{\fnms{ Vadim O.  } \snm{Sokolov}\thanksref{addr2}%
\ead[label=e3]{vsokolov@gmu.edu}}

\runauthor{Wang et al.}

\address[addr1]{Booth School of Business, University of Chicago, Chicago, IL, \printead{e1}, 
    \printead*{e2}.
}

\address[addr2]{Volgenau School of Engineering,  George Mason University, Fairfax, VA,
    \printead{e3}.
}

\end{aug}

\begin{abstract}
\noindent  Deep Learning (DL)  methods have emerged as one of the most powerful tools for  functional approximation and prediction.  While the representation properties of DL have been well studied, uncertainty quantification remains challenging and largely unexplored. Data augmentation techniques are a natural approach to provide uncertainty  quantification and to incorporate  stochastic Monte Carlo search into stochastic gradient descent (SGD) methods. The purpose of our paper is to show that training DL architectures with data augmentation leads to efficiency gains. We use the theory of scale mixtures of normals to derive data augmentation strategies for deep learning. This allows variants of the expectation-maximization and MCMC algorithms to be brought to bear on these high dimensional nonlinear deep learning models. To demonstrate our methodology, we develop  data augmentation algorithms  for a variety of commonly used activation functions: logit, ReLU, leaky ReLU and SVM. Our methodology is compared to traditional stochastic gradient descent with  back-propagation. Our optimization procedure leads to a version of iteratively re-weighted least squares and can be implemented at scale with accelerated linear algebra methods  providing substantial improvement in speed. We illustrate our methodology on a number of standard datasets.  Finally, we conclude with directions for future research.
\end{abstract}


\end{frontmatter}

\section{Introduction}

Deep neural networks (DNNs) have become a central tool for Artificial Intelligence (AI) applications such as, image processing (ImageNet, \citet{krizhevsky2012imagenet}), object recognition (ResNet, \citet{he2016deep}) and game intelligence (AlphaGoZero, \citet{silver2016mastering}). The approximability \citep{poggio2017and,bauer2017deep} and rate of convergence of  deep learning, either in the frequentist fashion \citep{schmidt2017nonparametric} or  from a Bayesian predictive point of view \citep{polson2018posterior, wang2020uncertainty},  have been well-explored and understood. \citet{fan2021selective} provides  a selective overview of deep learning. However, training deep learners is challenging due to the high dimensional search space and the non-convex objective function. Deep neural networks have also suffered from issues such as local traps, miscalibration and overfitting. Various efforts have been made to improve the generalization performance and many of  their roots lie in Bayesian modeling. For example, Dropout \citep{wager2013dropout} is commonly used  and  can be viewed as a deterministic ridge $\ell_2$ regularization. Sparsity structure via spike-and-slab priors \citep{polson2018posterior} on weights helps DNNs adapt to smoothness and avoid overfitting.  \citet{rezende2014stochastic} propose stochastic back-propagation through the use  of latent Gaussian variables.  

In this paper, following the spirit of  hierarchical Bayesian modeling, we develop data augmentation strategies for deep learning with a complete data likelihood function equivalent to weighted least squares regression. By using the theory of mean-variance mixtures of Gaussians, our latent variable representation brings all of the conditionally linear model theory to deep learning. For example, it allows for the straightforward specification of uncertainty at each layer of deep learning and for a wide range of regularization penalties.  Our method applies to commonly used activation functions such as ReLU, leaky ReLU, logit (see also \citet{gan2015learning}), and provides a general framework for training and inference in DNNs. It inherits the advantages and disadvantages of data augmentation schemes. For \textit{approximation} methods like Expectation-Maximization (EM) and Minorize-Maximization (MM), they are stable as they increase the objective but can be slow in the neighborhood of the maximum point even with acceleration methods such as Nesterov acceleration available and the performance is highly dependent on the properties of the objective function. Stochastic \textit{exploratory} methods like MCMC have the main advantage of addressing uncertainty quantification (UQ) and are stable in the sense they require no tuning.  Hyper-parameter estimation is immediately available using traditional Bayesian methods. DA augments the objective function with  extra hidden units which allow for efficient step size selection for the gradient descent search. In some of the applications, data augmentation methods can be formulated in terms of complete data sufficient statistics, a considerable advantage when dealing with large datasets where most of the computational expense comes from repeatedly iterating over the data. By combing the MCMC methods with the J-copies trick \citep{jacquier2007mcmc}, we can move faster towards posterior mode and avoid local maxima. Traditional methods for training deep learning models such as  stochastic gradient descent (SGD)  have none of the above advantages. We also note that we exploit the advantages of SGD and accelerated linear algebra methods when we implement our weighted least squares regression step.

Data augmentation strategies are commonplace in statistical algorithms and  accelerated convergence  \citep{nesterov1983method, green1984iteratively} is available. Our goal is to show similar efficiency improvements  for deep learning.  Our work builds on \citet{deng2019adaptive} who provide adaptive empirical Bayes methods. In particular, we show how to implement standard activation functions, including ReLU \citep{polson2018posterior}, logistic \citep{zhou2012beta, hernandez2015probabilistic} and SVM \citep{mallick2005bayesian} activation functions and provide specific data augmentation strategies and algorithms. The core subroutine of the resulting algorithms solves a least squares problem. Scalable linear algebra libraries such as Compute Unified Device Architecture (CUDA) and accelerated linear algebra (XLA) are available for implementation.   To  illustrate our approach, empirically  we experiment with two benchmark datasets using P\'{o}lya-Gamma data augmentation for logit activation functions. For the deep architecture embedded in our approach, we adopt deep ReLU networks. Deep  networks are able to achieve  the same level of approximation accuracy with exponentially fewer parameters for compositional functions \citep{mhaskar2017and}.  \citet{poggio2017and} further show how deep networks can avoid the curse of dimensionality. The ReLU function is favored due to its ability to avoid vanishing gradients and its expressibility and  inherent sparsity. Approximation properties of deep ReLU networks have been developed in \citet{montufar2014number}, \citet{telgarsky2017neural}, and \citet{liang2016deep}.  \citet{yarotsky2017error}  and \citet{schmidt2017nonparametric} show that deep ReLU networks can yield a rate-optimal approximation of smooth functions of an arbitrary order.  \citet{polson2018posterior} provide posterior rates of convergence for sparse deep learning.

There is  another active area of research that revives traditional statistical models with the computational power of DL \citep{bhadra2021merging}. Examples include   Gaussian Process models \citep{higdon2008computer, gramacy2008bayesian}, Generalized Linear Models (GLM) and Generalized  Linear Mixed Models (GLMM) \citep{tran2020bayesian}  and Partial Least Squares (PLS) \citep{polson2021deep}.  Our method benefits from the computation efficiency and flexibility of expression  of the deep neural network. In addition, our work builds on the sampling optimization literature \citep{pincus1968letter, pincus1970letter} which now uses MCMC methods.  Other examples include \citet{ma2018sampling} who study that sampling can be faster than optimization and \citet{neelakantan2015adding} showing that gradient noise can improve learning for very deep networks. \citet{gan2015learning} implements data augmentation inside learning deep sigmoid belief networks. \citet{neal2011mcmc} and \citet{chen2014stochastic} provide Hamitonian Monte Carlo (HMC) algorithms for MCMC.  \citet{duan2018scaling} proposes a family of calibrated data-augmentation algorithms to increase the effective sample size. 

The rest of our paper is outlined as follows. \Cref{sec:B-DL} provides the general setting of deep neural networks and shows how DA can be integrated into deep learning using the duality between Bayesian simulation and optimization.  \Cref{sec:DA} describes our data augmentation (DA) schemes and two approaches to implement them. \Cref{sec:app} provides  applications to Gaussian regression, support vector machines and logistic regression using P\'{o}lya-Gamma augmentation \citep{polson2013bayesian}. \Cref{sec:simu} provides the experiments of DA on both regression and classification problems. \Cref{sec:discussion} concludes with directions for future research.

\section{Bayesian Deep Learning}\label{sec:B-DL}
In deep learning we wish to recover a multivariate predictive map $f_{\btheta}(\cdot)$ denoted by 
\[\by=f_{\btheta}(\bx),\]
where $\by=(y_1, \ldots, y_n)', y_i \in \R$ denotes a univariate output and $\bx=(\x_1,\ldots, \x_n)'$, $\x_i\in \R^p$ a high-dimensional set of inputs.  Using training data of input-output pairs $\{y_i, \bx_i \}_{i=1}^n$ that generalizes well out-of-sample, the goal is to provide a predictive rule  for a new input variable $ \bx_\star$
\[
y_\star = f_{\hat\btheta}( \bx_\star ),
\]
where $\hat \btheta$ is estimated from training  data  typically  using SGD.  The interest in deep learners lies in their ability to perform better than the additive rule for those interpolation or prediction problems. Other statistical alternatives include Gaussian processes but they often have difficulty in handling higher dimensions. 

Deep learners  use compositions \citep{kolmogorov1957representation, vitushkin1964proof} of ridge functions rather than additive functions that are commonplace in statistical applications. With $L\in \mathbb{N}$ we denote the number of hidden layers and with $p_l\in \mathbb{N}$ the number of neurons at the $l^{th}$ layer. Setting $p_{L+1}=p, p_0=p_1=1$, we denote with $\mathbf{p}=(p_0,p_1, \ldots, p_{L+1})\in \mathbb{N}^{L+2}$ the vector of neuron counts for the entire network.  
Imagine composing $L$ layers, a deep predictor then takes the form
\begin{equation}\label{eq:DL_map}
\by=f_{\btheta}(\bx)=(  f_{W_0,b_0} \circ f_{W_1,b_1} \circ \cdots \circ f_{W_L,b_L}) (\bx),
\end{equation}
where $b_l\in \R^{p_l}$ is  a shift vector, $W_l \in \R^{p_{l-1}\times p_l}$ is a weight matrix that links neurons between $(l-1)^{th}$ and $l^{th}$ layers and  $f_{W_l,b_l}(x) =f_l( W_l x+b_l)$ is a semi-affine function. We denote with $\btheta=\{(W_0,b_0),(W_1,b_1),\ldots, (W_L,b_L)\}$ as the stacked parameters. We can rewrite the compositions in \eqref{eq:DL_map} with a set of latent variables  $Z=(Z_1, Z_2,$ $ \ldots, Z_L)'$ as
\begin{equation}\label{eq:DL}
\begin{split}
\y&=f_0(Z_1 W_0+b_0), \\
Z_{l}&=f_{l}( Z_{l+1} W_{l}+b_{l}), \quad l=1, \ldots, L, \\
Z_{L+1}&=\x,
\end{split}
\end{equation}
where $Z_l \in \R^{n\times p_l}$ is the matrix of hidden nodes in $l$-th layer.  We only consider the case $p=1$ and $Z_1\in \R^n$ in our work. We provide discussion on extensions to cases $p>1$ for some of our applications in \Cref{sec:app}.

\subsection{Bayesian Simulation and Regularization Duality }

The problem of deep learning regularization \citep{polson2017deep} is to find a set of parameters  $\btheta$ which minimizes a combination of a negative log-likelihood $\ell(\y, f_{\btheta}(\x))$ and  a penalty function $\phi(\btheta)$ defined by
\begin{equation}\label{eq:optim_obj}
\hat \btheta \defeq \arg \min_{\btheta}  \sum_{i=1}^n \ell(y_i, f_{\btheta} (\x_i)) +\lambda \sum_{j=1}^{\#{\btheta}} \phi(\theta_j),
\end{equation}
where $\lambda$ controls regularization and $\#{\btheta}$ denotes the number of parameters in $\btheta$.

When the function $f_{\btheta}(\x)$  is a deep learner defined as \eqref{eq:DL_map}, we can specify different  amount of penalty $\lambda_l$ and  form of regularization function $\phi_l(\cdot)$   for each layer. Then the objective function can be written as 
\begin{equation}\label{eq:obj_2}
\hat \btheta =\arg\min_{\btheta} \frac{1}{n}\sum_{i=1}^n \ell(y_i, f_{\btheta}(x_i))+\sum_{l=0}^{L} \lambda_l \phi_l(W_l, b_l).
\end{equation}
Commonly used regularization techniques for deep learners include $L^2$(weight decay), spike-and-slab regularization \citep{polson2018posterior} and dropout \citep{wager2013dropout}, which can also be viewed as a variant of $L^2$-regularization.

As such the optimization problem  in \eqref{eq:obj_2}  of training a deep learner  $f_{\btheta}(\cdot)$ involves a highly nonlinear objective function. Stochastic gradient descent (SGD)  is a popular tool based on  back-propagation (a.k.a. the chain rule), but it often suffers from local traps and overfitting   due  to the non-convex nature of the problem.  We propose data augmentation techniques  which can be seamlessly applied in this context  and provide efficiency gains.
This is achieved via the hierarchical duality between optimization with regularization and finding the  \textit{maximum a posteriori} (MAP)  estimate \citep{polson2011data},  as  described in the following lemma.
\begin{lemma} \label{lem:duality}
  The regularization problem
  \begin{equation*}
   \hat \btheta= \arg \min_{\btheta} \, \left\{ \frac{1}{n}\sum_{i=1}^n \ell(y_i, f_{\btheta}(x_i))+\sum_{l=0}^{L} \lambda_l \phi_l(W_l, b_l)\right\}
   \end{equation*}
  is equivalent to finding the the Bayesian MAP estimator defined by
  \[
   \arg\max_{\btheta} \,  p({\btheta}|\by) =\arg \max_{\btheta} \,\exp\left\{-\frac{1}{n}\sum_{i=1}^n \ell(y_i, f_{\btheta}(x_i))-\sum_{l=0}^{L} \lambda_l \phi_l(W_l, b_l)\right\}, \]
which corresponds to the  mode of a posterior distribution characterized as
 \begin{gather*}
p(\btheta\mid  \y) =  p(\y \mid \btheta) p(\btheta)/p(\y),\\
p(\by|\btheta)\propto \exp\{-\sum_{i=1}^n \ell\big(y_i, f_{\btheta} (\x_i)\big)\},\quad p(\btheta)\propto \exp\{-\sum_{l=0}^{L} \lambda_l \phi_l(W_l, b_l)\}.
\end{gather*}
Here $p(\btheta)$ can be interpreted as a prior probability distribution and the log-prior as the regularization penalty.
  \end{lemma}

\subsection{A  Stochastic Top  Layer}

By exploiting the duality from \Cref{lem:duality},  we  wish to  use a Bayesian framework to add stochastic layers -- so as to  fully  account  for the uncertainty in estimating the  predictive rule $f_{\btheta}(\cdot)$. Thus, we convert the sequence  of composite functions in the deep learner  specified  in \eqref{eq:DL} to  a stochastic version given by 
\begin{equation}\label{eq:DL_stochastic}
\begin{split}
\y \mid Z_1 &\sim p(\y \mid Z_1), \\
Z_{l} &\sim N(f_{l}(W_{l} Z_{l+1}+b_{l}), \tau_{l}^2), \quad l=1,2, \ldots, L, \\
Z_{L+1}&=\x.
\end{split}
\end{equation}
Now the hidden variables $Z=(Z_1, \ldots, Z_L)'$  can be viewed as data augmentation variables and hence will also allow the contribution of fast scalable algorithms for  inference  and  prediction.

For the ease of  computation, we only  replace the top layer of the  DNN with  a stochastic layer. We denote network structure below the top layer with  $\B=\{(W_1, b_1), \ldots, (W_L,b_L)\}$, and the network structure can be rewritten as
\begin{align*}
\by=f_0( Z_1 W_0 +b_0),\quad  Z_1 = f_{\B}(\bx),
\end{align*}
where the function $f_0(Z_1 W_0+b_0)$ is the top layer structure and function $f_{\B}(\bx)$ is the network architecture below the top layer. Considering the objective function in \eqref{eq:obj_2}, we implement the solutions with a two-step iterative search. At iteration $t$, we have
\begin{enumerate}[noitemsep]
\vspace{-0.1in}
\item DA-update for the top layer ${W_0, b_0}$  as the MAP estimator of the distribution
\begin{align}
p(W_0, b_0 \mid Z_1^{(t)}, \y)&\propto p(\y, Z_1^{(t)}\mid W_0, b_0 )p(W_0, b_0) \label{eq:top} \\
&\propto \exp\left\{- \frac{1}{n} 
\sum_{i=1}^n \ell(y_i, f_{\btheta}(x_i)\mid \B^{(t)})+\lambda_0 \phi_0(W_0, b_0)\right\}\nonumber
\end{align}
\item SGD-update for the deep architecture $\B$
\begin{align*}
\B^{(t+1)}&=\arg \min_{\B}  \frac{1}{n} 
\sum_{i=1}^n \ell\big(y_i, f_{\btheta}(x_i)\mid (W_0, b_0)^{(t+1)}\big)+\sum_{l=1}^L\lambda_l \phi_l(W_l, b_l)\\
&=\arg \min_{\B}  \frac{1}{n} 
\sum_{i=1}^n \ell\big(Z_1^{(t)}, f_{\B}(x_i)\big)+\sum_{l=1}^L\lambda_l \phi_l(W_l, b_l).
\end{align*}
\item Sample $Z_1^{(t+1)}$ from a normal distribution $\mN(\mu_z^{(t)}, \sigma_z^{(t)})$ where $\mu_z^{(t)}$ and $\sigma_z^{(t)}$ are determined jointly by $\{\theta^{(t)}, \x, \y\}$.
\vspace{-0.1in}
\end{enumerate}

The main contribution of our work comes from two aspects: (1) we  update top layer weights $\{W_0, b_0\}$ conditional on $\B$ as in \eqref{eq:top}, which is also equivalent to conditioning on $Z_1$, with data augmentation techniques as later shown in  \Cref{sec:DA}; (2) the latent variables $Z_1$ is sampled from a normal distribution rather than optimized by gradient descent methods. $Z_1$ serves as a bridge that connects a  weighted $L^2$-norm model $f_0$ and a deep learner $f_{\B}$.  Commonly used activation functions $\{f_l\}_{l=1}^L$ are linear affine functions, rectified linear units (ReLU), sigmoid, hyperbolic tangent (tanh), and etc.  We illustrate our methods with a deep ReLU network, i.e., $\{f_l\}_{l=1}^L$ are ReLU functions, due to its expressibility and  inherent sparsity. In the next section, we introduce our data augmentation strategies and show how the stochastic layers can be achieved via data augmentation.

\section{Data Augmentation  for Deep Learning}\label{sec:DA}

Data augmentation introduces a vector of auxiliary variables, denoted by  $\bomega=(\omega_1, \ldots, \omega_n)'$ with $\omega_i\in \R$, such that  the posterior can be written as
\[p(\btheta\mid \by)= E_{\bomega} \Big[  p(\btheta,  \bomega\mid \by) \Big],\] 
where the augmented auxiliary  distribution, $p(\btheta, \bomega  \mid \by)$ factorizes  nicely   into complete conditionals $p(\btheta \mid \bomega,  \by)$ and $p(\bomega\mid  \btheta,  \by)$. A crucial ingredient is that $p(\btheta\mid \bomega,  \by) $ is easily managed typically via conditional Gaussians. 

Data augmentation tricks allow us to express the likelihood  as an expectation of a weighted $L^2$-norm. Specifically, we write
\begin{align*}
\exp\Big\{-\ell\big(\y, f_{\btheta}(\x)\big)\Big\}&=E_\omega \Big\{\exp\Big(-Q\big( \y \mid f_{\btheta}(\x),\bomega \big)\Big)\Big\}\\
&=\int_0^\infty \exp\Big(-Q\big( \y\mid f_{\btheta}(\x),\bomega \big)\Big)p(\bomega)\d\bomega
\end{align*}
where $p(\bomega)$ is the prior on the auxiliary variables $\bomega=(\omega_1, \ldots, \omega_n)'$ and the function $Q\big(  \y \mid f_{\btheta}(\x), \bomega\big)$ is designed to be a quadratic form, given the data augmentation variables $\bomega$. The function $f_{\btheta}(\x) = (f_0 \circ \cdots \circ f_L)(\x)$ is a deep learner.

 \Cref{tab:identities} shows that standard activation functions such as ReLU, logit, lasso and check can be expressed in the form of  \eqref{eq:quadratic}. Commonly used activation functions for deep learning, with an appropriate stochastic assumptions for $w$ (for notation of simplicity, we derive the standard form for the single observation case) can be expressed as 
\begin{align*}
\exp(-\max(1-x,0))&=E_\omega \left\{\frac{1}{\sqrt{2\pi\omega}}\exp\Big(-\frac{1}{2\omega}(x-1-\omega)^2\Big)  \right\},  &\text{ where } & \omega \sim \gig(1,0,0), \\
\exp(-\log(1+e^{x}))&=E_\omega \left\{\exp(-\frac{1}{2}\omega x^2 ) \right\}, &\text{ where }& \omega \sim \pg(1,0), \\
\exp(-|x|)&=E_\omega \left\{\frac{1}{\sqrt{2\pi\omega}}\exp\Big(-\frac{1}{2\omega}x^2\Big) \right\}, &\text{ where }& \omega \sim \mE \big(\frac{1}{2}\big).
\end{align*}
Here $\mathcal{GIG}$ denotes the Generalized Inverse Gaussian distribution,  $\pg$ represents the P\'{o}lya Gamma  distribution \citep{polson2013bayesian}, and $\mE$ represents the exponential distribution.

\begin{table}[!ht]
\centering
\scalebox{0.7}{
\begin{tabular}{lll}
\toprule
  $l(W,b)$ & $Q(W,b,\omega)$ & $p(\omega)$\\
\midrule
\vspace{0.08in}
ReLU: ${\max(1-z_i,0)}$ & $\displaystyle \int_0^\infty \frac{1}{\sqrt{2\pi c\lambda}}\exp
\left\{-\frac{(x+a\lambda)^2}{2c\lambda}\right\} \d \lambda  =\frac{1}{a} \exp \left ( - \frac{ 2\max ( ax , 0 )}{c} \right )$   & $\gig(1,0,0)$ \\
\vspace{0.08in}
Logit: $\log(1+e^{z_i})$ &  $\displaystyle \frac{1}{2^b} e^{(a-b/2) \psi}\int_0^\infty e^{-\omega \psi^2/2} p(\omega) \d \omega = \frac{(e^\psi)^a}{(1+e^\psi)^b}$ &$\pg(b,0)$\\
\vspace{0.08in}
Lasso: $|\frac{z_i}{\sigma}|$ &$\displaystyle \int_0^\infty \frac{1}{\sqrt{2\pi c\lambda}}  \exp\left\{-\frac{x^2}{2c\lambda}\right\} e^{ - \frac{1}{2} \lambda } \d \lambda  = \frac{1}{c}\exp \left ( - \frac{ | x |}{c}  \right ) $ & $\mE(\frac{1}{2})$\\
\vspace{0.08inx}
Check: $|z_i|+(2\tau-1)z_i$   & $\displaystyle  \int_0^\infty \frac{1}{\sqrt{2\pi c\lambda}}  \exp\left\{-\frac{(x+(2\tau-1)\lambda)^2}{2c^2\lambda}\right\}  e^{ -  2 \tau ( 1 - \tau ) \lambda }  \d \lambda  = 
 \frac{1}{c} \exp \left ( - \frac{2}{c} \rho_\tau (  x ) \right )$ & $\gig(1,0, 2\sqrt{\tau-\tau^2})$\\
\bottomrule
\end{tabular}}
\caption{Data Augmentation Strategies. Here $ \rho_\tau ( x ) = \frac{1}{2} | x | + \left ( \tau - \frac{1}{2} \right ) x$ is the check function.}\label{tab:identities} 
\end{table}

Using the data augmentation strategies, the objectives are represented as   mixtures of Gaussians. DA can perform such an optimization with only the use of a sequence of iteratively re-weighted $L^2$-norms. This allows us to use XLA techniques to accelerate the training. 

\begin{remark}
The log-posterior is optimized given the training data, $\{y_i, \x_i\}_{i=1}^n$. Deep learning possesses the key property that $\nabla_{\btheta} \log p(\by|\theta,\bx)$ is computationally inexpensive to evaluate using tensor methods for very complicated architectures and fast implementation on large datasets. One caveat is that the posterior is highly multi-modal and providing good hyperparameter tuning can be expensive. This is clearly a fruitful area of research for state-of-the-art stochastic MCMC algorithms to provide more efficient algorithms. For shallow architectures, the alternating direction method of multipliers (ADMM) is an efficient solution to the optimization problem.
\end{remark}

Similarly we can represent  the regularization penalty  $\exp(-\lambda \phi(\btheta))$ in data augmentation form. Hence, we can then replace  the  optimization  problem in \eqref{eq:obj_2}  with
\begin{equation}\label{eq:quadratic}
\hat \btheta \defeq \arg\max_{\btheta}  E_{\bomega}\Big[\exp\Big(-\frac{1}{n}\sum_{i=1}^n Q\big(y_i \mid f_{\btheta}(\x_i), \bomega\big)-\sum_{l=0}^{L} \lambda_l \phi_l(W_l, b_l) \Big) \Big],
\end{equation}
using the duality in \Cref{lem:duality}.

There are two approaches to Monte Carlo optimization which could handle our data augmentation \citep{geyer1996estimation}, missing data methods like Expectation-Maximization (EM) algorithms or stochastic search methods like Markov Chain Monte Carlo (MCMC). The first approach is based on a probabilistic \textit{approximation} of the objective function \eqref{eq:quadratic} and is less concerned with exploring $\bTheta$. The second type is more \textit{exploratory}  which aims to optimize the objective function by visiting the entire range of $\bTheta$ and is less tied to the properties of the function.
 
 For EM algorithms, we consider constructing a surrogate optimization problem which has the same solution to \eqref{eq:quadratic} \citep{lange2000optimization}. Specifically, we define  a new objective function as
\[
H(\btheta)=\E_{\bomega\mid \btheta}\Big[\exp\Big(-\frac{1}{n}\sum_{i=1}^n  Q\big(y_i \mid f_{\btheta}(\x_i), \bomega\big)-\sum_{l=0}^L\lambda_l \phi_l(W_l, b_l) \Big)\Big],
\]
which is a concave function to be maximized. A natural choice of the surrogate function can be constructed  using Jensen's inequality as
\[
 G\big(\btheta \mid \btheta^{(t)} \big) = - \E_{\bomega\mid \btheta^{(t)}} \left[ \frac{1}{n}\sum_{i=1}^n  Q\big( y_i \mid  f_{\btheta}(\x_i), \bomega \big) +\sum_{l=0}^L \lambda_l \phi_l(W_l, b_l) \right],
\]
where each  $\omega_i$ is drawn from conditional distribution $p(\omega_i \mid \btheta)\propto p(\omega_i, \btheta)$ and the minorization is satisfied as 
\[
\log H(\btheta)\geq G\big(\btheta \mid \btheta^{(t)} \big).
\]
Maximizing $G\big(\btheta \mid \btheta^{(t)} \big)$ with respect to $\btheta$ drives $H(\btheta)$ uphill. The ascent property of the EM algorithm relies on the nonnegativity of the Kullback-Leibler divergence of two conditional probability densities \citep{hunter2004tutorial,lange2013mm}. The EM algorithm enjoys the numerical stability as it steadily increases the likelihood  without wildly overshooting or undershooting. It simplifies the optimization problem  by (1) avoiding large matrix inversion; (2) linearizing the objective function; (3) separating the variables of the optimization problem \citep{lange2013optimization}. In  \Cref{sec:logit} we show how P\'{o}lya-Gamma augmentation leads to an EM algorithm for logistic regression.

The exploratory alternative to solve \eqref{eq:quadratic} is stochastic search methods such as MCMC. The data augmentation strategies enable us to sample from the joint posterior 
\begin{align*}
p(\btheta\mid \y)&\propto \exp\Big(-\frac{1}{n}\sum_{i=1}^n \ell\big(y_i, f_{\btheta}(\x_i)\big)-\sum_{l=0}^L\lambda_l \phi_l(W_l, b_l) \Big) \\
&= E_{\bomega}\Big[\exp\Big(-\frac{1}{n}\sum_{i=1}^n  Q\big(y_i \mid f_{\btheta}(\x_i), \bomega\big)-\sum_{l=0}^L\lambda_l \phi_l(W_l, b_l) \Big)\Big]\\
& = \int_0^\infty  \exp \Big(-\frac{1}{n}\sum_{i=1}^n  Q(y_i \mid f_{\btheta}(\x_i), \bomega \big) \Big)p(\bomega) p(\btheta) \d\bomega
\end{align*}
where the prior is related to the regularization penalty, via  $p(\btheta)\propto \exp\big(-\sum_{l=0}^L\lambda_l \phi_l(W_l, b_l) \big)$. 

Hence, we can provide an MCMC algorithm in the augmented space $(\btheta, \bomega)$ and simulate from the joint posterior distribution, denoted by $p(\btheta,  \bomega\mid \y)$, namely
\[p(\btheta, \bomega\mid \by) \propto \exp\Big(-Q(\y \mid f_{\btheta}(\x),\bomega)\Big)p(\btheta)p(\bomega).
\]
A sequence can be simulated using  MCMC  Gibbs conditionals,
\begin{align*}
p\big(\btheta^{(t)} \mid \bomega^{(t)}, \y\big) & \propto \exp\Big(-Q(\y\mid f_{\btheta}(\x), \bomega^{(t)})\Big)p(\btheta), \\
p\big(\bomega^{(t+1)} \mid \btheta^{(t)},\y\big) &\propto \exp\Big(-Q( \y\mid f_{\btheta^{(t)}}(\x),\bomega)\Big)p(\bomega).
\end{align*}
Then we recover stochastic draws $\btheta^{(t)}\sim p(\btheta\mid \by )$ from the  marginal posterior. These draws can be used in  prediction to  account for  \textit{predictive uncertainty}, namely
\begin{equation}\label{eq:mcmc_pred}
p\big(y_\star\mid f(\x_\star)\big) =\int  p\big(y_\star\mid \btheta, f_{\btheta} (\x_\star)\big) p(\btheta\mid \y)\d \btheta \approx \frac{1}{T}\sum_{t=1}^T p\big(y_\star \mid \btheta^{(t)}, f_{\btheta^{(t)}} (\x_\star) \big).
\end{equation}

As $Q( \y \mid f_{\btheta}(\x),\bomega)$ is conditionally quadratic, the update step for $\btheta \mid \bomega, \y$ can be achieved using SGD or a weighted $L^2$-norm -- the weights $\bomega$ are adaptive and provide an automatic choice of the learning rate, thus avoiding backtracking which can be computationally expensive. And the performance of MCMC search is less tied to the statistical properties (i.e. convexity or concavity) of the objective function. We provide examples of how Gaussian regression and SVMs can be implements in \Cref{sec:gr} and \Cref{sec:svm}.
   
\subsection{MCMC with J-copies}
The MCMC methods offer a full description of the objective function \eqref{eq:quadratic} over the entire space $\bTheta$. Inspired by the simulated annealing algorithm \citep{metropolis1953equation}, we introduce a scaling factor $J$ to allow for faster moves on the surface of \eqref{eq:quadratic} to maximize. It also helps avoiding the trapping attraction of local maxima. In addition, the corresponding posterior is connected to the Boltzmann distribution, whose  density is prescribed by the energy potential $f(\theta)$ and temperature $J$ as
\begin{equation}\label{eq:boltzmann}
\pi_J (\btheta) = \exp \left \{ -J f(\btheta) \right \} / Z_J \; \; {\rm for} \; \; \btheta \in \bTheta
\end{equation}
where $ Z_J = \int_{\bTheta} \exp \left \{ - J f(\btheta) \right \} d \btheta $ is an appropriate normalizing constant.

\begin{figure}[!ht]
\centering
\includegraphics[width=0.7\textwidth, height=0.6\textwidth]{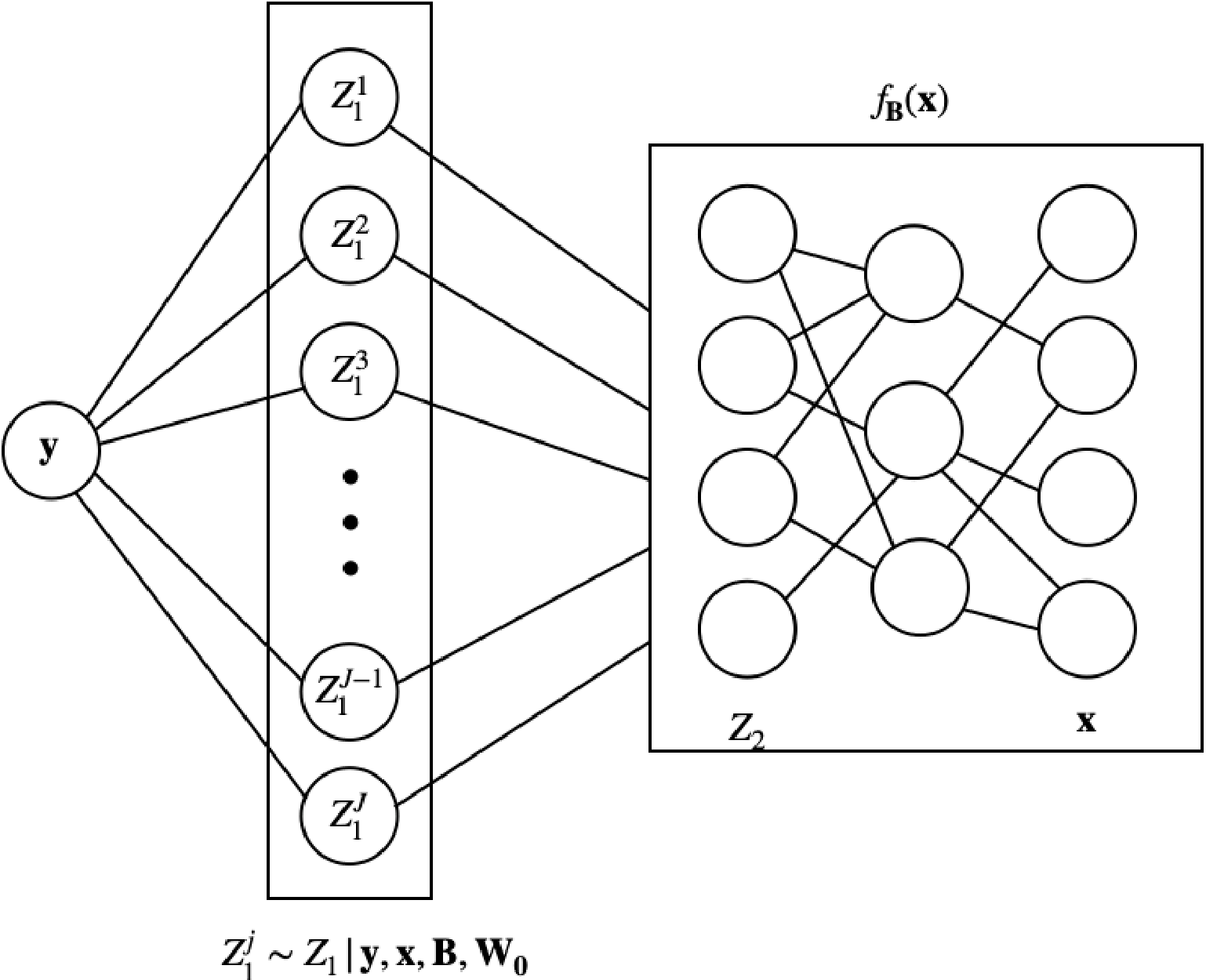}
\vspace{-0.1in}
\caption{$J$-copies Network Architecture}\label{fig:arch}
\vspace{-0.1in}
\end{figure}

To simulate the posterior mode without evaluating the likelihood directly  \citep{jacquier2007mcmc}, we sample $J$ independent copies of hidden variable $Z_1$ .  Denoted the copies with $Z_1^1, \ldots, Z_1^J$,  we sample them simultaneously and independently from the posterior distribution 
\[Z_1^j|\btheta, \bx, \by \iid \mN(\mu_z, \sigma_z^2), \quad j=1, \ldots, J,\]
where $\mu_z, \sigma_z$ are determined by $\{\x, \y, \btheta\}$. And we stack the J copies as
\begin{equation}\label{eq:stacked}
\by^{(S)}=\left[\begin{array}{c}
\by\\
\by\\
\by\\
\vdots\\
\by
\end{array}\right], \quad Z_1^{(S)}=\left[\begin{array}{c}
Z_1^1\\
Z_1^2\\
Z_1^3\\
\vdots\\
Z_1^J
\end{array}\right],\quad {f_{\B}}(\bx^{(S)})=\left[\begin{array}{c}
f_{\B}(\bx)\\
f_{\B}(\bx)\\
f_{\B}(\bx)\\
\vdots\\
f_{\B}(\bx)
\end{array}\right]
\end{equation}
where $\by^{(S)}$, $Z_1^{(S)}$ and ${f_{\B}}(\bx^{(S)})$ are $(n\times J)$-dimensional vectors. We use $Z_1^{(S)}$ to amplify the information in $\by$, which is especially useful in the finite sample problems. \Cref{fig:arch} illustrates our network architecture.

With the stacked system, the joint distribution of the parameters $\btheta$ and the augmented hidden variables $Z_1^{(S)}$ given data $\by, \bx$ can be written as
\[
\pi_J(\btheta,  Z_1^{(S)}\mid \bx, \by)\propto  \prod_{j=1}^J p(\by \mid \btheta,  Z_1^j)p(Z_1^j\mid \btheta, \x, \by) p(\theta).
\]
Hence,  the marginal joint posterior
 \[
 p(\btheta \mid \bx, \by) = \int \pi_J(\btheta, Z_1^{(S)} \mid \bx, \by) \d  Z_1^{(S)}
 \]  
concentrates on the density proportional to  $p( \bx,\by \mid \btheta)^Jp(\btheta)$ and  provides us with  a simulation solution to finding the MAP estimator \citep{pincus1968letter, pincus1970letter}.

Another alternative to simulate from the posterior mode is Hamiltonian Monte Carlo \citep{neal2011mcmc}, which is a modification of Metropolis-Hastings (MH) sampler. Adding an additional momentum variable $\bm{\nu}$ to the Boltzmann distribution in \eqref{eq:boltzmann}, and generating draws from joint distribution
\[
\pi_J(\btheta, \bomega ) \propto \exp \left(-J f(\btheta) - (1/2) {\bm{\nu}}^TM^{-1} \bm{\nu}\right),
\]
where $M$ is a mass matrix. \citet{chen2014stochastic} adopt this approach in a deep learning setting.

\subsection{Connection to Diffusion Theory}

An alternative to the MCMC algorithm can be derived from diffusion theory \citep{phillips1996bayesian}. For example, we can approximate the random walk Metropolis-Hastings algorithm with the Langevin diffusion $L_t$ defined by the stochastic differential equation $dL_t= dB_t+\frac{1}{2}\nabla \log f(L_t) dt$, where $B_t$ is the standard Brownian motion. More specifically, let $d:=\abs{\btheta}$, we write the random walk like transition as
\[  
\btheta^{(t+1)}=\btheta^{(t)}+\frac{\sigma^2}{2}\nabla \log f(\btheta^{(t)})+\sigma \bepsilon_t,
\]
where $\bepsilon_t \sim \mN_d(0, I_d)$ and $\sigma^2$ corresponds to the discretization size.

This can also be derived by taking a second-order approximation of $\log(f)$, namely
\begin{align*}
\log f(\btheta^{(t+1)})=&\log f(\btheta^{(t)})+\big(\btheta^{(t+1)}-\btheta^{(t)}\big)' \nabla \log f(\btheta^{(t)})\\
&-\frac{1}{2}\big(\btheta^{(t+1)}-\btheta^{(t)}\big)'  H(\btheta^{(t)})\big(\btheta^{(t+1)}-\btheta^{(t)}\big),
\end{align*}
where $H(\btheta^{(t)})=-\nabla^2 \log f(\btheta^{(t)})$ is the Hessian matrix. By taking exponential transformation on both sides, the random walk type approximation to $f(\btheta^{(t+1)})$ is
\begin{align*}
f(\btheta^{(t+1)}) & \propto \exp\Big\{ \big(\btheta^{(t+1)}-\btheta^{(t)}\big)'\nabla \log f(\btheta^{(t)})-\frac{1}{2} \big(\btheta^{(t+1)}-\btheta^{(t)}\big)'H(\btheta^{(t)})  \big(\btheta^{(t+1)}-\btheta^{(t)}\big)\Big\}\\
&\propto \exp\Big\{  -\frac{1}{2}\big(\btheta^{(t+1)}-\widetilde\btheta^{(t)}\big)' H(\btheta^{(t)})\big(\btheta^{(t+1)}-\widetilde\btheta^{(t)}\big)\Big\}.
\end{align*}
where $\widetilde \btheta^{(t)}=\btheta^{(t)}+H^{-1}(\btheta^{(t)})\nabla \log f(\btheta^{(t)})$. If we simplify this approximation by replacing $H(\btheta^{(t)})$ with $\sigma^{-2} I_p$, the Taylor approximation leads to updating step as
\[
\btheta^{(t+1)}=\btheta^{(t)}+\sigma^2\nabla \log f(\btheta^{(t)})+\sigma \bepsilon_t.
\]
\citet{roberts1998optimal} give further discussion on the choice of $\sigma$ that would yield an  acceptance rate of 0.574 to achieve optimal convergence rate.


\citet{mandt2017stochastic} show that SGD can be interpreted as a multivariate Ornstein-Uhlenbeck process
\[
d \btheta^{(t)}=-\eta A \btheta^{(t)} dt +\eta \sqrt\frac{C}{ S}  d W^{(t)},
\]
here $\eta$ is the constant learning rate, $A$ is the symmetric Hessian matrix at the optimum and $\frac{C}{S}$ is the covariance of the mini-batch (of size $S$) gradient noise, which is assumed to be approximately constant near the local optimum of the loss. They also provide results on discrete-time dynamics on other Stochastic Gradient MCMC algorithms, such as Stochastic Gradient Langevin dynamics (SGLD) by \citet{welling2011bayesian}  and Stochastic Gradient Fisher Scoring by \citet{ahn2012bayesian}.

Combing their results and the Langevin dynamics of MCMC algorithms, we can write the approximation of our DA-DL updating scheme as
\begin{align*}
\left[\begin{array}{c}
W_0\\
b_0
\end{array}
\right]^{(t+1)}&=\left[\begin{array}{c}
W_0\\
b_0
\end{array}
\right]^{(t)}+\sigma^2\nabla\log f_0(Z_1^{(t)}W_0^{(t)}+b_0^{(t)})+\sigma \bepsilon_{0t},\\
\B^{(t+1)}&=\B^{(t)}-\eta\nabla^2 f_{\B^*}(\bx)\B^{(t)}+\frac{C}{\sqrt S}\eta \bepsilon_{\B t}.
\end{align*}

Similar adaptive dynamics are also observed in other methods. \citet{geman1986diffusions} show the convergence of the annealing process using Langevin equations.  Slice sampling \citep{neal2003slice} adaptively chooses the step size based on  the local properties of the density function. By constructing local quadratic approximations, it could adapt to the dependencies between variables. \citet{murray2010elliptical} further propose elliptical slice sampling that operates on the ellipse of states.

\section{Applications}\label{sec:app}
To illustrate our methodology, we provide three examples: (1) a standard Gaussian regression model with squared loss; (2) a binary classification model under the support vector machine framework; (3) a logistic regression model paired with a P\'{o}lya mixing distribution. For the Gaussian regression and SVM models, we implement with $J$-copies stacking strategy to provide full posterior modes. 

Before diving into the examples, we introduce the notations we use throughout this section. We continue to denote the output with $\y=(y_1, \ldots, y_n)', y_i\in \R$, the input with $\x=(\x_1, \ldots, \x_n)', \x_i \in \R^p $, the latent variable of the top layer  with $Z_1=(z_{1,1}, \ldots, z_{1,n})', z_{1,i}\in \R$ and the stacked version as in \eqref{eq:stacked}. We introduce stochastic noises $\bepsilon_{0}=(\epsilon_{0,1}, \ldots, \epsilon_{0,n})'$ in the top layer and $\bepsilon_{z}=(\epsilon_{z,1}, \ldots, \epsilon_{z,n})'$ in the second layer, where $\epsilon_{0,i}\iid\mN(0, \tau_0^2)$ and $\epsilon_{z,i}\iid \mN(0, \tau_z^2)$. The scale parameters $\tau_0$ and $\tau_z$ are pre-specified and determine the level of randomness or uncertainty for the DA-update and SGD-update respectively. We use $\eta$ to denote the learning rate used in the SGD updates and $T$ is number of  training epochs. We use $\norm{\cdot}$ to denote $\ell_2$-norm such that $\norm{\y}=\sqrt{\sum_{i=1}^n y_i^2}$ and the matrix-type norm as $\norm{\y}_\Sigma=\sqrt{ y^T \Sigma y}$.

Our models differ from standard deep learning models and some newly proposed Bayesian approaches in the adoption of stochastic noises $\bepsilon_0$ and $\bepsilon_z$. It distinguishes our model from other deterministic neural networks. By letting $\bepsilon_z$ follow a spiky distribution that puts most of its mass around zero,  we can control the estimation approximating to posterior mode instead of posterior mean. The randomness allows us to adopt a stacked system and make the best use of data especially when the dataset is small. 

\subsection{Gaussian Regression}\label{sec:gr}
We consider the regression model as
\begin{align*}
y_i&=z_{1,i}W_0+b_0+\epsilon_{0,i}, \, &\text{ where }& \, y_i  \in (-\infty,\infty),\, \epsilon_{0,i} \overset{i.i.d}{\sim} \mN(0, \tau_0^2),\\
z_{1,i} & = f_{\B}(x_i)+\epsilon_{z,i}, \, &\text{ where }& \,\epsilon_{z,i} \overset{i.i.d}{\sim} \mN(0, \tau_z^2).
\end{align*}
The posterior updates are given by
\begin{align}
\hat W_0 &= \Cov(Z_1,\by)/\Var(Z_1)\label{eq:W0_gauss},\\
\hat b_0 &= \bar{\by}-W_0\bar{Z_1} \label{eq:b0},\\
p(Z_1\mid \y, \x, \btheta) &= C_z \exp\left\{-\frac{1}{2\tau_0^2} \norm{\by-Z_1W_0-b_0}^2-\frac{1}{2\tau_z^2}\norm{Z_1- f_{\B}(\bx)}^2 \right\}, \nonumber
\end{align}
where $\bar{\by}=\frac{1}{n}\sum_{i=1}^n y_i$ and $C_z$ is a normalizing constant. The latent variable $Z_1$ is drawn from following normal distribution $Z_{1} \sim \mN(\mu_Z, \sigma_Z^2)$ 
with the mean and variance specified as
\begin{equation}\label{eq:Z0_gauss}
\mu_Z=\frac{\tau_z^2 W_0(\by- b_0)+\tau_0^2f_{\B}(\bx)}{ W_0^2\tau_z^2+\tau_0^2}, \quad \sigma_Z^2=\frac{\tau_0^2\tau_z^2}{ W_0^2\tau_z^2+\tau_0^2}.
\end{equation}
The $J$ copies of $Z_1$ are simulated and stacked as
\begin{align*}
Z_1^j \iid  \mN(\mu_Z, \sigma_Z^2), \quad Z_1^{(S)}=(Z_1^1, \ldots, Z_1^J)'.
\end{align*} 
The updating scheme  for this Gaussian regression is summarized in  \Cref{alg:gr}.

\begin{algorithm}[ht]
\caption{Data Augmentation with $J$-copies for Gaussian Regression (DA-GR)}\label{alg:gr}
\begin{algorithmic}[1]
\State Initialize $ \B^{(0)},W_0^{(0)}, b_0^{(0)}$
\For{epoch $t=1, \ldots, T$}

\State{Update the weights in the top layer with $\{\by^{(S)}, Z_1^{(t,S)}\}$}

$W_0^{(t)}=\Cov(Z_1^{(t,S)},\by^{(S)})/\Var(Z_1^{(t,S)}) $ 

$b_0^{(t)}=\bar\by^{(S)}-W_0^{(t)}\bar{Z_1}^{(t,S)}$

\State{Update the deep learner $f_{\B}$ with $\{Z_1^{(t,S)}, \bx^{(S)}$\}}

 $\B^{(t)}=\B^{(t-1)}-\eta\nabla f_{\B^{(t-1)}}\big(\bx^{(S)}\mid Z_1^{(t,S)}\big) $ \Comment{SGD}

\State{Update $Z_1^{(S)}$ jointly  from deep learner $f_{\B}$ and sampling layer $f_0$}

 ${Z_1}^{j, (t+1)}\mid W_0^{(t)},b_0^{(t)},\by,  f_{\B^{(t)}}(\bx) \iid \mN\big(\mu_z^{(t)},{\sigma_z^{(t)}}^2\big), \, j=1,\ldots, J$
\EndFor\\
\Return $\hat \by=  W_0^{(T)} f_{\B^{(T)}}(\bx) + b_0^{(T)}$
\end{algorithmic}
\end{algorithm}

The model can also be generalized to multivariate $y$. Let $y_i$ be a $q$-dimension vector,  we denote each dimension as $y_{ik}, k=1,\ldots, q$, and the model is written as 
\begin{align*}
y_{ik}&=z_{1,i}W_{0k}+b_{0k}+\epsilon_{0,ik}, \, &\text{ where }& \, y_{ik}  \in (-\infty,\infty),\, \epsilon_{0,ik} \iid \mN(0, \tau_0^2),\\
z_{1,i} & = f_{\B}(x_i)+\epsilon_{z,i}, \, &\text{ where }& \,\epsilon_{z,i} \iid \mN(0, \tau_z^2),
\end{align*}
where $W_0=(W_{01}, \ldots, W_{0q})'$ is now a q-dimensional vector with $W_{0k}$ computed similarly to \eqref{eq:W0_gauss}, $b_0=(b_{01}, \ldots, b_{0q})'$ is also q-dimensional with $b_{0k}$ calculated as  \eqref{eq:b0}. The posterior update for $Z_1$ becomes
\[
p(Z_1 \mid \by, \bx,\btheta) = C_z \exp\left\{-\frac{1}{2\tau_0^2}\sum_{k=1}^q \norm{\by_k-Z_1W_{0k}-b_{0k}}^2-\frac{1}{2\tau_z^2}\norm{Z_1- f_{\B}(\bx)}^2 \right\},
\]
which is  a multivariate normal distribution with the  mean and variance as
\[
\mu_Z=\frac{\tau_z^2 \sum_{k=1}^q W_{0k}(\by_k- b_{0k})+\tau_0^2f_{\B}(\bx)}{ \tau_z^2\sum_{k=1}^q W_{0k}^2+\tau_0^2}, \sigma_Z^2=\frac{\tau_0^2\tau_z^2}{\tau_z^2 \sum_{j=k}^q W_{0k}^2+\tau_0^2}.
\]

\subsection{Support Vector Machines (SVMs)}\label{sec:svm}

Support vector machines require data augmentation for rectified linear unit (ReLU) activation functions.  \citet{polson2011data} and \citet{mallick2005bayesian} write the support vector machine model as
\[
\by=Z_1W_0+ \blambda+\sqrt{\blambda} \bepsilon_0, \,\text{ where } \blambda\sim p(\blambda),
\]
where $p(\blambda)$ follows a flat uniform prior. The augmentation variable $\blambda=(\lambda_1, \ldots, \lambda_n)'$  can be regarded as slacks admitting fuzzy boundaries between classes.

By incorporating the augmentation variable $\blambda$, the ReLU deep learning model can be written as
\begin{align*} 
 y_i&= z_{1,i}W_0+\lambda_{i}+\sqrt{\lambda_{i}}\epsilon_{0,i}, \, \text{ where } \, y_i \in \{-1, 1\}, \epsilon_{0,i} \overset{i.i.d}{\sim} \mN(0,\tau_0^2),\\
z_{1,i} &= f_{\B}(\bx_i)+\epsilon_{z,i}, \text{ where } \, \epsilon_{z,i} \overset{i.i.d}{\sim} \mN(0, \tau_z^2).
\end{align*}
 From a probabilistic perspective, the likelihood function for the output $\by$ is given by
\begin{align*}
p (y_i \mid W_0, z_{1,i}) & \propto \exp \left\{-\frac{2}{\tau_0^2}\max (1-y_i z_{1,i}W_0,0)\right\}\\
& \propto \int_0^\infty \frac{1}{\tau_0 \sqrt{2\pi \lambda_{i}}} \exp \left( -\frac{1}{2\tau_0^2}\frac{(1+\lambda_{i}-y_i z_{1,i}W_0)^2}{\lambda_{i}}   \right)\d \lambda_{i}.
\end{align*}
Derived from this augmented likelihood function, the conditional updates are
\begin{align*}
   W_0 \mid \by, Z_1, \blambda &\propto \left[ \prod_{i=1}^n \frac{1}{\tau_0\sqrt{\lambda_{i}}} \right]\left[ \exp\left\{-\frac{1}{2\tau_0^2} \sum_{i=1}^n\frac{(1+\lambda_{i}-y_iz_{1,i}W_0)^2}{\lambda_{i}} \right\} \right]\\
     Z_1 \mid \by, \bx ,W_0, \B &\propto \exp\left\{-\frac{1}{2\tau_0^2}\norm{\by-Z_1W_0}^2_{\Lambda^{-1}}-\frac{1}{2\tau_z^2}\norm{Z_1-f_{\B}(\bx)}^2\right\} 
\end{align*}
where $\Lambda=\diag(\lambda_{1},\ldots, \lambda_{n})$ is the diagonal matrix of the augmentation variables.

In order to generate the latent variables, we use conditional Gibbs sampling as
\begin{align}
 \lambda_{i}^{-1}\mid W_0, y_i, z_{1,i} &\sim \mathcal{IG}(\abs{1-y_iz_{1,i}W_0}^{-1},{\tau_0^{-2}})\label{eq:lambda}\\
W_0 \mid \by, Z_1,\lambda & \sim \mN(\mu_w, \sigma_w^2)\label{eq:W0_bin}\\
Z_1\mid \by, \bx,W_0, \B & \sim \mN(\mu_z, \sigma_z^2)\label{eq:Z0_bin}
\end{align}
with the means and variances given by
\[
\mu_w=\frac{\sum_{i=1}^n y_i z_{1,i} \frac{1+\lambda_i}{\lambda_i}}{\tau_0^2\sum_{i=1}^n \frac{y_i^2z_{1,i}^2}{\lambda_i}}, 
\sigma_w^2=\frac{1}{\tau_0^2\sum_{i=1}^n \frac{y_i^2z_{1,i}^2}{\lambda_i}}, 
\mu_z=\frac{W_0\tau_z^2 \by+\tau_0^2f_{\B}(\bx)\Lambda\mathbf{1}}{W_0\tau_z^2+\tau_0^2\Lambda\mathbf{1}},
\sigma_z^2=\frac{\tau_0^2\tau_z^2\Lambda\mathbf{1}}{W_0^2\tau_z^2+\tau_0^2\Lambda\mathbf{1}},
\]
where $\mathcal{IG}$ denotes the Inverse Gaussian distribution and $\mathbf{1}=(1, \ldots, 1)'$ is a n-dimensional unit vector.

The $J$-copies strategy can also be adopted here. $Z_1^j$ and $\blambda^j$ needs to be sampled independently for $j=1, \ldots, J$. \Cref{alg:svm} summarizes the updating scheme with $J$-copies for SVMs. 

\begin{algorithm}[ht]
\caption{Data Augmentation with $J$-copies for SVM (DA-SVM)}\label{alg:svm}
\begin{algorithmic}[1]
\State Initialize $ \B^{(0)},W_0^{(0)}, \blambda^{(0)}$
\For{epoch $t=1, \ldots, T$}

\State{Update the weights and slack variables in the top layer with $\{\by^{(S)}, Z_1^{(t,S)}\}$}

$ \{\blambda^{(t, S)}\}^{-1}\mid W_0^{(t-1)}, \by^{(S)}, Z_1^{(t, S)} \sim \mathcal{IG}(|1-\by^{(S)} Z_1^{(t,S)}W_0^{(t-1)}|^{-1},\frac{1}{\tau_0^2}) $

$W_0^{(t)}\mid \by^{(S)},Z_1^{(t, S)},\blambda^{(t, S)}  \sim \mN(\mu_\omega^{(t)}, {\sigma_\omega^{(t)}}^2) $

\State{Update the deep learner $f_{\B}$ with $\{Z_1^{(t, S)}, \bx^{(S)}\}$}

 $\B^{(t)}=\B^{(t-1)}-\eta\nabla f_{\B^{(t-1)}}\big(\bx^{(S)}\mid Z_1^{(t,S)}\big) $ \Comment{SGD}


\State{Update $Z_1^{(S)}$ jointly  from the deep learner $f_{\B}$ and the sampling layer $W_0$}

$Z_1^{j, (t+1)}\mid W_0^{(t)},\blambda^{j, (t)}, \by, f_{\B^{(t)}}(\bx) \iid \mN(\mu_z^{(t)},{\sigma_z^{(t)}}^2), \, j=1,\ldots, J$ 
\EndFor\\

\Return $\hat y= \left\{\begin{array}{l} 
1, \text{ if } W_0^{(T)} f_{\B^{(T)}}(\bx)>0\\
-1, \text{ otherwise. }
\end{array}\right. $
\end{algorithmic}
\end{algorithm}

\subsection{Logistic Regression}\label{sec:logit}
The aim of this example is to show how EM algorithm can be implemented via a weighted $L^2$-norm in deep learning. Adopting the logistic regression model from \citet{polson2013data}, we focus on the penalization of $W_0$, with parameter optimization given by
\[\hat W_0 =\arg \min_{W_0} \left[ \frac{1}{n}\sum_{i=1}^n \log\Big(1+\exp\big(-y_i f^{DL}_{\B}(\x_i)W_0\big)\Big)+\phi(W_{0}\mid \tau) \right],\]
The outcomes $y_i$ are coded as $\pm 1$, and $\tau$ is assumed fixed.

 For likelihood function $\ell$ and  regularization penalty $\phi$, we assume
\begin{align}
p(y_i \mid \sigma)&\propto \int_{0}^\infty \frac{\sqrt \omega_i}{\sqrt{2\pi}\sigma}\exp\left\{
-\frac{\omega_i}{2\sigma^2} \big(y_if_{\B}(\x_i) W_0-\frac{1}{2\omega_i}\big)^2\right\}  p(\omega_i) \d\omega_j,\\
p(W_{0}\mid \tau)&=\int_0^\infty\frac{\sqrt{\lambda}}{\sqrt{2\pi}\tau} \exp\left\{-\frac{\lambda}{2\tau^2} (W_{0}-\mu_W-\kappa_W\lambda^{-1})^2\right\}p(\lambda)\d\lambda,
\end{align}
where $\mu_W, \kappa_W$ are pre-specified terms controlling the prior of the penalty term and $\lambda$ is endowed with a P\'{o}lya distribution prior $P(\lambda)$. Let $\omega_i^{-1}$ have a P\'{o}lya distribution with $\alpha=1,\kappa=1/2$, the following three updates will generate a sequence of estimates that converges to a stationary point of posterior
\begin{align*}
\displaystyle
& W_0^{(t+1)}=(\tau^{-2}\Lambda^{(t)}+\bx_*^T \Omega^{(t)}\bx_*)^{-1}(\frac{1}{2}\bx^T_*\mathbf{1}),\\
&  \omega_i^{(t+1)}=\frac{1}{z_i^{(t+1)}}\left( \frac{e^{z_i^{(t+1)}}}{1+e^{z_i^{(t+1)}}}-\frac{1}{2}\right),  \lambda^{(t+1)} =\frac{\kappa_W+\tau^2 \phi'(W_{0}^{(t)}\mid \tau)}{W_{0}^{(t)}-\mu_W},
\end{align*}
where $z_i^{(t)}=y_i z_{1,i}^T{W_0^{(t)}}=y_i \text{logit}(\hat y_i^{t})$, $\bx_*$ is a matrix with rows $x_i^*=y_iz_{1,i}$,  $\Omega=\text{diag}(\omega_1, \ldots, \omega_n)$ and $\Lambda=\text{diag}(\lambda_1, \ldots, \lambda_p)$ are diagonal matrices. $\bx_*$ can be written as $\bx_*=\text{diag}(\by) Z_1$, $\phi'(\cdot)$ denotes the derivative of standard normal density function.

In the  non-penalized case, with $\lambda_i=0$ for every $i$, the updates can be simplified as weighted least squares
\begin{align*}
\displaystyle
& W_0^{(t+1)}=({Z_1^{(t)}}^T \text{diag}(\by) \Omega^{(t)}\text{diag}(\by)Z_1^{(t)})^{-1}(\frac{1}{2} \by^T{Z_1^{(t)}} ),\\
&  \omega_i^{(t+1)}=\frac{1}{z_i^{(t+1)}}\left( \frac{e^{z_i^{(t+1)}}}{1+e^{z_i^{(t+1)}}}-\frac{1}{2}\right).
\end{align*}
We focus on the non-penalized binary classification case and \Cref{alg:logit} summarizes our approach. Further generalizations are available. For example, a ridge-regression penalty, along with the generalized double-pareto prior \citep{armagan2013generalized} can be implemented by adding a sample-wise $L^2$-regularizer. A multinomial generalization of this model can be found in \citet{polson2013data}.

\begin{algorithm}[ht]
\caption{Data Augmentation for Logistic Regression (DA-logit)}\label{alg:logit}
\begin{algorithmic}[1]
\State Initialize $W_0^{(0)}, b_0^{(0)} \B^{(0)}$
\For{epoch $t=1, \ldots, T$}

\State{ Retrieve the input and output of the top layer}

$Z_1^{(t)}=f_{{\B}^{(t-1)}}(\bx)$ \Comment{input}

$\by^{(t)}=\text{sigmoid}(W_0^{(t-1)}Z_1^{(t)}+b_0^{(t-1)})$\Comment{output}

\State{ Calculate the sample-wise weights}

 $\bz^{(t)}=\by\cdot \text{logit}( \by^{(t)})$ \Comment{transformed responses}
 
$\bomega^{(t)}= \frac{1}{\bz^{(t)}} (\text{sigmoid}(\bz^{(t)})-\frac{1}{2}) $ \Comment{weights}

\State{ Update the entire deep learner $f_{\btheta}$ with $\{\by,\bx\}$}

${\btheta}^{(t)}={\btheta}^{(t-1)}-\eta\nabla f_{{\btheta}^{(t-1)}}(\bx\mid\by, \text{sample\_weights} = \bomega^{(t)}) $ \Comment{SGD}
\EndFor
\\

\Return $\hat \by= \left\{\begin{array}{l} 
1, \text{ if } f_{ {\btheta}^{(T)}}(\bx)>\frac{1}{2}\\
-1, \text{ otherwise. }
\end{array}\right. $
\end{algorithmic}
\end{algorithm}

\section{Experiments}\label{sec:simu}
We illustrate the performance of our methods on both synthetic and real datasets, compared to the deep ReLU networks without the data augmentation layer. We refer to the latter as DL in our results. We denote the data augmented gaussian regression in \Cref{alg:gr} as DA-GR, the SVM implementation in \Cref{alg:svm} as DA-SVM and the logistic regression in \Cref{alg:logit} as DA-logit. For appropriate comparison, we adopt the same network structures, such as the number of layers, the number of hidden nodes, and  regularizations like dropout rates, for DL and our methods.  The differences between our methods and DL are that (1) the top layer weights ${W_0, b_0}$ of DL are updated via SGD optimization, while the weights ${W_0, b_0}$ of our methods are updated via MCMC or EM; (2) for binary classification, DA-logit and DL adopt a sigmoid activation function in the top layer to produce a binary output, while DA-SVM uses a linear function in the top layer and the augmented sampling layer transforms the continuous value into a binary output. For all experiments, the datasets are partitioned into 70\% training  and 30\% testing randomly. For the optimization we use a modification of the SGD algorithm, the Adaptive moment estimation (Adam, \citet{kingma2014adam}) algorithm. The Adam algorithm combines the estimate of the stochastic gradient with the earlier estimate of the gradient, and scales this using an estimate of the second moment of the unit-level gradient. We have also explored  RMSprop \citep{tieleman2012rmsprop} optimizer and we observe similar decreases in regression or classification errors.

To illustrate how the choice of $J$ could affect the speed of convergence, we include different implementations of DA-GR and DA-SVM with $J=2, 5, 10$. We have explored different sampling noise variance $\tau_0, \tau_Z$, but the choices, in general, do not affect the results significantly. 

\subsection{Friedman Data}\label{sec:friedman}
The benchmark  \citep{friedman1991multivariate}  setup uses a  regression of the form
\[y_i= 10\sin(\pi x_{i1}x_{i2}) + 20(x_{i3} -0.5)^2 + 10x_{i4} + 5x_{i5}+\epsilon_i, \quad \text{with}\quad \epsilon_i\sim\mathcal{N}(0,\sigma^2),\]
where $\x_i= (x_{i1}, x_{i1}, \ldots, x_{ip})$ and only the first 5 covariates are predictive of $y_i$. We run the experiments with $n=100, 1\,000$ and $p=10, 50 , 100, 1\,000$ to explore the performance in both low dimensional and high dimensional scenarios. We implement both one-layer ($L=1$) and two-layer $(L=2)$ ReLU networks with 64 hidden units in each layer. For DA-GR model, we let $\tau_0=0.1, \tau_z=1$.  The experiments are repeated 50 times with different random seeds.

\begin{figure}[!ht]
\begin{center}
\includegraphics[height=0.95\textwidth,angle=270]{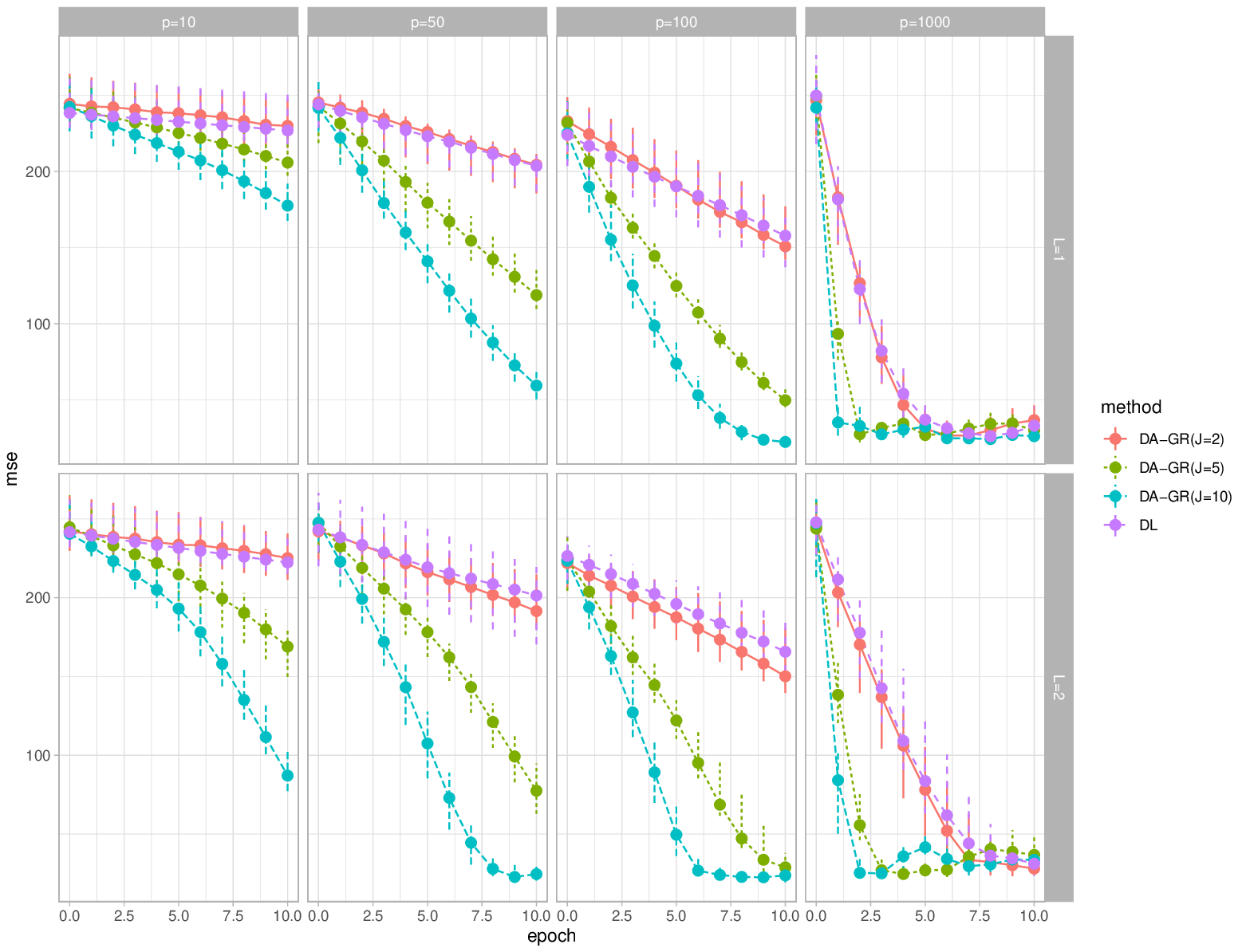}
\vspace{-0.15in}
\end{center}
\caption{Quartiles of out-of-sample MSEs under the Friedman Setup. We explore cases where  $n=1\,000$ and $p=10, 50, 100, 1\,000$. The tests are repeated $50$ times. The medians of out-of-sample MSEs after  training for 1 to 10 epochs are plotted with lines and the vertical bars mark the 25 \% and 75\% quantiles of the MSEs. DA-GR refers to  DA Gaussian regression shown in \Cref{alg:gr} and DL stands for the ReLU networks without the data-augmentation layer.} \label{fig:friedman}
\vspace{-0.15in}
\end{figure}

\Cref{fig:friedman} reports the three quartiles of the out-of-sample squared errors (MSEs). The top row is the performance of the one-layer networks and the bottom row is the performance of the two-layer networks. The two-layer networks perform better and converge faster. For DA-GR, when $J=5$ or $J=10$, it converges significantly faster and the prediction errors are also smaller. When $J=2$, the performance of DA-GR is relatively similar to the deep learning model with only SGD updates. This is due to the fact that DA-GR with J-copies learns the posterior mode which is equivalent to the minimization point of the objective function, and it concentrates on the mode faster when $J$ becomes larger.

\begin{figure}[!ht]
\begin{center}
\includegraphics[height=0.95\textwidth,angle=270]{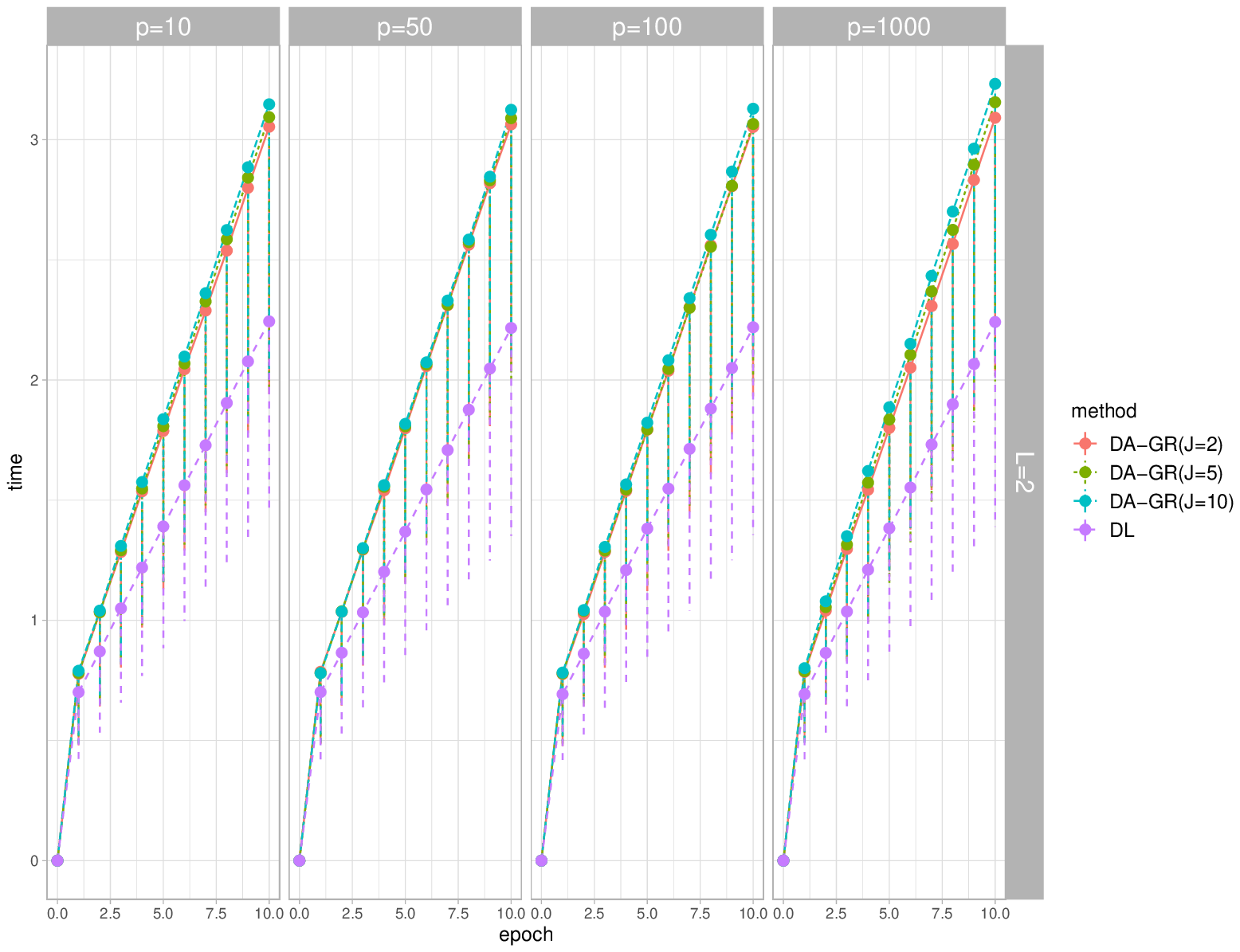}
\end{center}
\vspace{-0.15in}
\caption{Computation Time under the Friedman setup. The setups are $n=1\,000$ and $p=10, 50, 100, 1\,000$.  The averaged time (over 50 repetitions) for computing 1 to 10 epochs is plotted with lines and the vertical bars mark the 25\% and 75\% quantiles of the computation time collected.  We only include one figure of computation time comparison here since the scale is relatively the same for all cases.}\label{fig:comp}
\vspace{-0.15in}
\end{figure}

The computation costs of DA are higher as shown in \Cref{fig:comp}. This is not entirely unexpected since we introduce sampling steps. When $J$ increases, the computation costs also increase slightly. Given the improvement in convergence speed and prediction errors, our data augmentation strategies are still worthwhile even with some extra computation costs. In addition, for each epoch, we can draw the sample-wise posteriors in parallel and the gap between the computation time can be further  mitigated.

\subsection{Boston Housing Data}
Another classical regression benchmark dataset is the Boston Housing dataset\footnote{https://archive.ics.uci.edu/ml/machine-learning-databases/housing/}, see, for example, \citet{hernandez2015probabilistic}.
The data contains $n=506$ observations with 13 features. 
 To show the robustness of DA, we repeat the experiment 20 times with different training subsets. We adopt the ReLU networks with one hidden layer of 64 units and set the dropout rate to be 0.5. For the DA-GR model, we let $\tau_0=0.1, \tau_Z=1$.

\begin{figure}[!ht]
\centering
\includegraphics[height=0.95\textwidth,  angle=270]{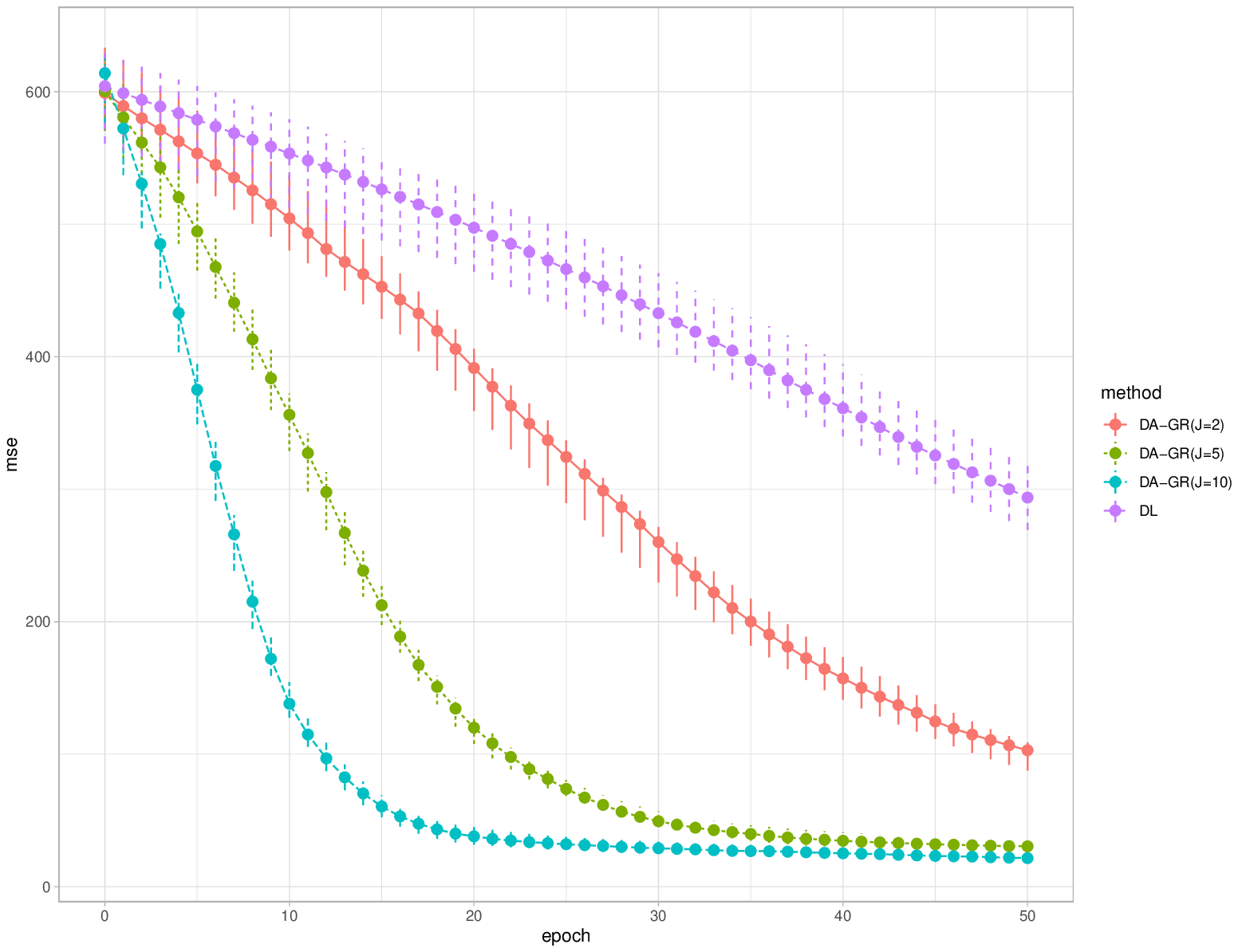}
\caption{Out-of-sample MSEs for the Boston Housing dataset. The experiment is repeated 20 times with different training subsampling. The medians of MSEs after training for 1 to 50 epochs are provided, with the vertical bars marking the 25\% and 75\% quantiles of the errors. DA-GR refers to the data augmentation strategy in \Cref{alg:gr} and DL stands for the ReLU networks without the data-augmentation layer.}\label{fig:boston}
\vspace{-0.1in}
\end{figure}

\Cref{fig:boston} shows the prediction errors of all methods. DA-GR with $J=10$ performs significantly better than the others, in terms of both prediction errors and convergence rates. Meanwhile, DA-GR with $J=2$ behaves similarly to SGD at the beginning, but it converges significantly faster than SGD after a few epochs. This again, shows that with the J-copies strategy, our method helps the optimization converge at a faster speed, and injecting the noise helps the model generalize well out-of-sample.

\subsection{Wine Quality Data Set}
The Wine Quality Data Set \footnote{P. Cortez, A. Cerdeira, F. Almeida, T. Matos and J. Reis, `Wine Quality Data Set', UCI Machine Learning Repository.} contains 4\,898 observations with 11 features. The output  wine rating  is an integer variable ranging from 0 to 10 (the observed range in the data is from 3 to 9). The frequency of each rating is reported in \Cref{tab:wine}.
\begin{table}[!ht]
\centering
\begin{tabular}{l | c c c c c c c }
\toprule
rating & 3 &   4 &   5 &   6 &   7 &   8 &   9  \\
\midrule
frequency & 20 & 163 & 1457 & 2198 & 880 & 175 &   5\\
\bottomrule
\end{tabular}
\caption{Frequencies of Different Wine Ratings}\label{tab:wine}
\vspace{-0.1in}
\end{table}

The most frequent ratings are 5 and 6. Since we focus on binary classification problems, we provide two types of classifications, both of which have relatively balanced categories: (1) wine with a rating of 5 or 6 (Test 1);  (2) wine with a rating of $\leq 5$ or $>5$ (Test 2). We use the same network architectures adopted in Friedman's example with $\tau_0=\tau_z=0.1$.

\begin{figure}[!ht]
\includegraphics[height=0.95\textwidth, angle=270]{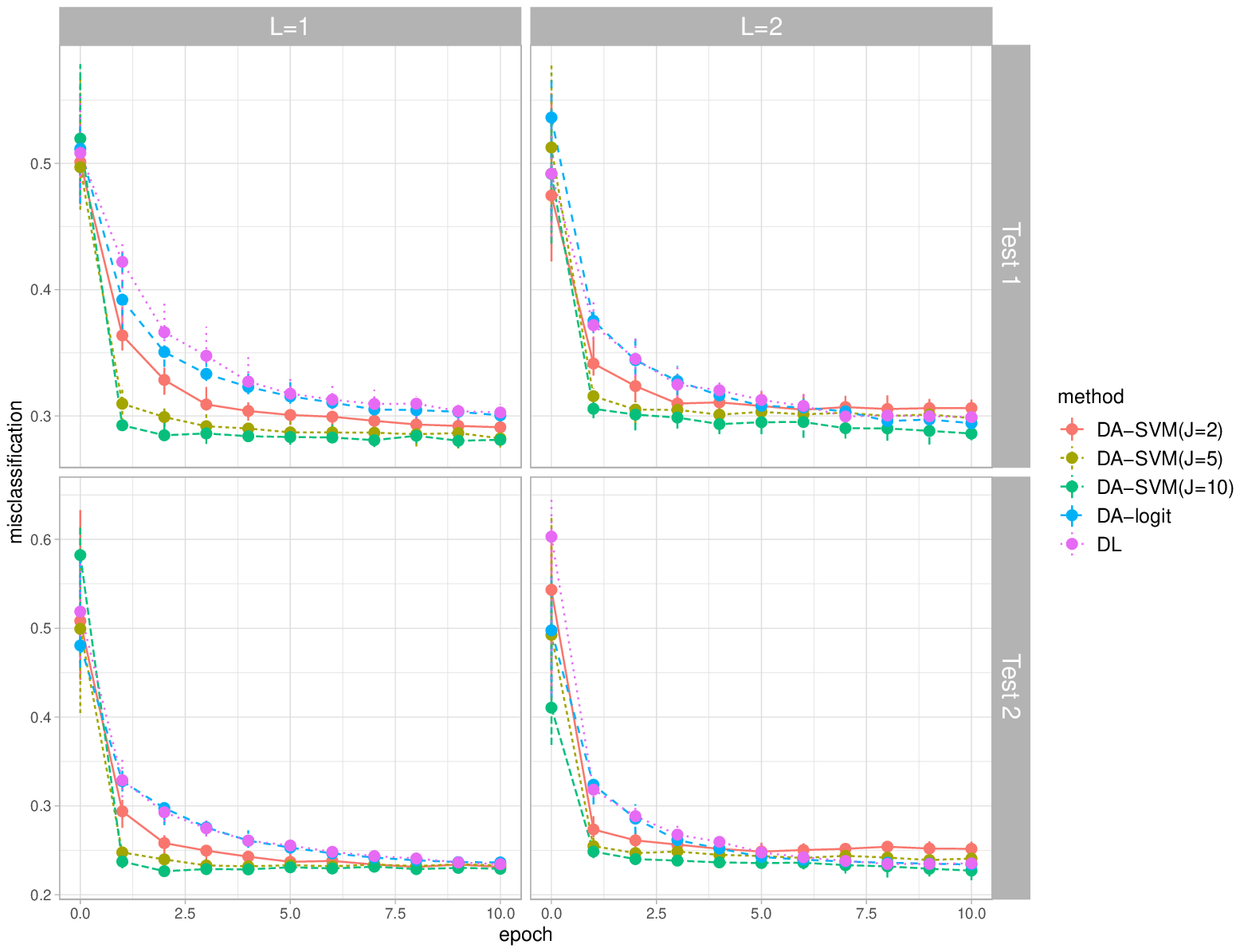}
\caption{Binary Classifications on the Wine Quality dataset. Two types of binary classifications are considered here. The experiment is repeated 20 times with different training subsampling. We compare the misclassification rates of DA-SVM in \Cref{alg:svm} with $J=2, 5, 10$, DA-logit in \Cref{alg:logit} and the ReLU networks without the data augmentation layer (DL), after training for 1 to 10 epochs.}\label{fig:wine}
\vspace{-0.1in}
\end{figure}

 \Cref{fig:wine} provides results for the two types of binary classifications. In both cases, DA-SVM performs better than DA-logit and DL.  The advantage of large $J$ is still significant and helps converge especially in the early phase. DA-logit outperforms DL in Test 1 when the network is shallow (L=1), while in other cases performs similarly to DL. 

\subsection{Airbnb Data Set}
 The Airbnb Kaggle competition\footnote{https://www.kaggle.com/c/airbnb-recruiting-new-user-bookings} provides a more challenging application with  21\,3451 observations in total, and classified by destination into 12 classes: 10 most popular countries, other and  no destination found (NDF), where \textit{other} corresponds to any other country which is not among the top 10 and NDF corresponds to situations that no booking was made.   The countries are denoted with their standard codes, as `AU' for Australia, `CA' for Canada, `DE' for Germany, `ES' for Spain, `FR' for France, `UK' for United Kingdom, `IT' for Italy, `NL' for Netherlands, `PT' for Portugal, `US' for United States. \Cref{tab:dest} reports the percentage of each class. We follow the preprocessing steps in \citet{polson2017deep}. The list of variables contains information from the sessions records (number of sessions, summary statistics of action types, device types and session duration), and user tables such as gender, language, affiliate provider etc. All categorical variables are converted to binary dummies, which leads to 661 features in total. For the neural network architecture, we use a two-layer ReLU network with 64 hidden units on each layer and set the dropout rate to be 0.3. For the SVM model, we let $\tau_0=\tau_z=0.1$.

\begin{table}[!ht]
\footnotesize
\centering
\begin{tabular}{l | *{12}{c} }
\toprule
 & AU& CA & DE & ES& FR& UK &IT& NDF& NL& PT& US& other \\
\midrule
\% obs &  0.25 &  0.67 &  0.50 & 1.05 & 2.35 & 1.09 & 1.33 & 58.35 & 0.36 & 
 0.10 & 29.22 &  4.73 \\
\bottomrule
\end{tabular}
\caption{Percentage of Each Class (\#obs = 21\,3451)}\label{tab:dest}
\end{table}

Our goal is to  test the  binary classification models on this dataset.  We consider two types of binary responses, both of which have relatively balanced amounts of observations in each category.
\begin{enumerate}
\vspace{-0.1in}
\item Spain (1.05\%) vs United Kingdom(1.09\%) 
\item United Kingdom (1.09\%) vs Italy (1.33\%)
\vspace{-0.1in}
\end{enumerate}

\begin{figure}[!ht]
\centering
\includegraphics[width=0.98\textwidth, height=0.4\textwidth]{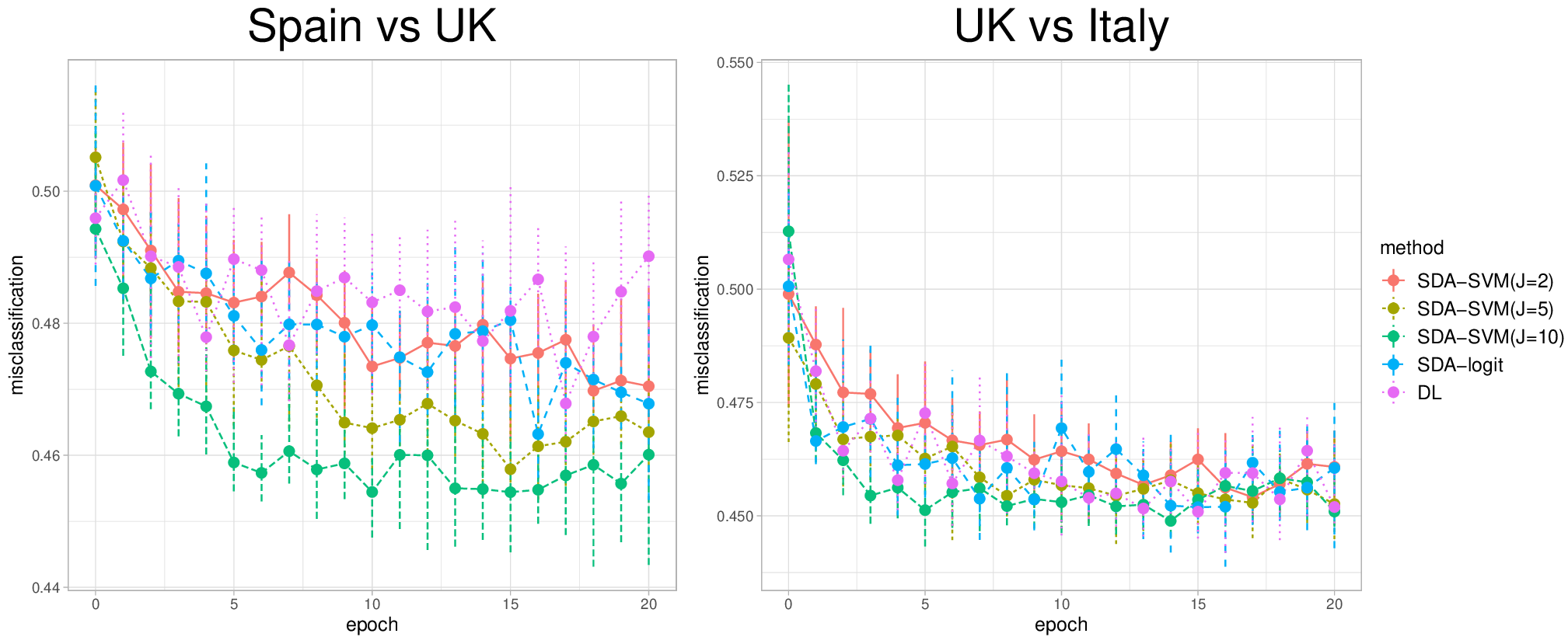}
\vspace{-0.15in}
\caption{Binary Classifications on the Airbnb Booking Dataset. Two types of binary classifications are considered here. The experiment is repeated 20 times with different training subsampling. We compare the misclassification rates of DA-SVM in \Cref{alg:svm} with $J=2, 5, 10$, DA-logit in \Cref{alg:logit} and the ReLU networks without the data augmentation layer, after training for 1 to 20 epochs. }\label{fig:airbnb}
\vspace{-0.15in}
\end{figure}

\Cref{fig:airbnb} demonstrates the binary classifications for Spain versus UK and UK versus Italy. For both cases, the out-of-sample misclassification rates are not small  and the fluctuations over epochs are big, suggesting that a better model structure may be needed. However, we still observe that DA-SVM with $J=5$ or $J=10$ has  smaller classification errors over epochs and the out-of-sample errors decrease faster during earlier phase of training.


\subsection{Summary of Experiment Results}

From the above examples, we observe that DA-logit which is implemented under the EM principle does not show an obvious advantage over the vanilla neural network. It shows some improvements on the convergence speed when the network is shallow in the Wine Quality dataset case as in \Cref{fig:wine}. This could be partially due to the fact that we did not apply regularization on the DA layer for our logit implementation. More importantly, the performance of the EM algorithm is contingent on the statistical properties of the objective function. Although the surrogate function is constructed via only the top layer whose quadratic form ensures concavity, the property of the objective function as a whole becomes complicated when the deep network architecture is more complex. Since our method also inherits the negative side of EM and MM algorithms, convergence to the global maximum is not guaranteed in the absence of concavity.  However,  this observation could open the possibility of future research where we can combine the EM algorithms with shape-constrained neural networks \citep{gupta2020multidimensional}.

On the contrary, the MCMC methods with the J-copies strategy significantly improve the prediction errors and convergence speed of the neural networks for both regression and classification problems. And the advantages become more outstanding when $J$ is larger. The phenomenon suggests that the stochastic exploratory methods are preferable when the statistical property of the objective function is unknown or too complex. And the $J$-copies scheme largely relieves the problem of  being trapped into local modes.

One concern of using MCMC methods is the extra computation costs induced by the sampling steps. In our current version where $p_1=1$, the sample-wise sampling steps can be computed in parallel. If one wishes to introduce a higher dimension latent variable $Z_1$ such that $p_1>1$, the computation costs will increase as it may involve sampling from multivariate distributions. In that case, fast sampling implementation such as \citet{bhattacharya2016fast} is recommended to speed up the process.

\section{Discussion}\label{sec:discussion}
Various regularization methods have been deployed in neural networks to prevent overfitting, such as early stopping, weight decay, dropout \citep{hinton2012improving}, gradient noise \citep{neelakantan2015adding}.  Bayesian strategies tackle the regularization problem by proposing probability structures on the weights. We show that  data augmentation strategies are available for many standard activation functions (ReLU, SVM, logit) used in deep learning. 
 
Using MCMC  provides a natural stochastic search mechanism that avoids procedures such as back-tracking and provides full descriptions of the objective function over the entire range $\bTheta$. Training deep neural networks thus benefits from additional hidden stochastic augmentation units (a.k.a. data augmentation).  Uncertainty can be injected into the network through the probabilistic distributions on only one or two layers, permitting more variability of the network. When more data are observed, the level of uncertainty decreases as more information is learned and the network becomes more deterministic. We also exploit the duality between \textit{maximum a posteriori} estimation and optimization.  We provide a $J$-copies stacking scheme to speed up the convergence to posterior mode and avoid trapping attraction of the local modes. Concerning efficiency, DA provides a natural framework to convert the objective function into weighted least squares and  is straightforward to implement with the current deep learning training process. 

Our three motivational examples illustrated the advantages of data augmentation. Our work has the potential to be generalized to many other data augmentation schemes and different regularization priors. Probabilistic structures on more units and layers are also possible to allow for more uncertainty. 

Our DA-DL methods enjoy the benefits of both worlds. On one hand, with the data augmentation on top,  it is  robust to random weight initialization. Although we still need to specify the learning rates for the deep architecture, the top layer can learn adaptively and the entire network becomes less sensitive to the choice of learning rate. On the other hand, the fast SGD updates from the deep architecture largely alleviate the computation concerns compared to a fully Bayesian hierarchical model.

There are many directions to future research, including adding more sampling layers so the model could accommodate more randomness and flexibility, and using weighted Bayesian bootstrap \citep{newton2018weighted} to approximate the unweighted posteriors by assigning random weight to each observation and penalty. Uncertainty quantification for prediction is also possible. Although we focus on the training aspect of deep learning, one can collect posterior draws $\btheta^{(t)}$  from the MCMC procedure when the training process converges. Using \eqref{eq:mcmc_pred}, we can construct predictive intervals and conduct inference.

\bibliographystyle{ba}
\bibliography{dl-mcmc}
 
\end{document}